\documentclass{article}
\usepackage{spconf,amsmath,graphicx}

\usepackage{makecell,hhline,lscape,array,ifthen,colortbl,multirow,siunitx,xspace}
\usepackage{algorithm,algpseudocode,dsfont,wrapfig,xcolor,pifont,float,placeins,import,etoolbox,booktabs}
\usepackage[inkscapelatex=false]{svg}
\usepackage[utf8]{inputenc}
\usepackage[T1]{fontenc}
\usepackage{url,amsfonts,nicefrac,microtype}
\usepackage{hyperref}
\usepackage{enumitem}
\setlist{nosep, leftmargin=14pt}

\makeatletter
\DeclareRobustCommand\onedot{\futurelet\@let@token\@onedot}
\def\@onedot{\ifx\@let@token.\else.\null\fi\xspace}

\def\eg{\emph{e.g}\onedot} 
\def\ie{\emph{i.e}\onedot}

\def\wrt{w.r.t\onedot} 

\makeatother

\newcommand{\ResM}{\mbox{\textsc{ResMatching}}\xspace}
\newcommand{\DeMicFlow}{\mbox{\textsc{HazeMatching}}\xspace}
\newcommand{\InDI}{\mbox{\textsc{InDI}}\xspace}

\newcommand{\RCAN}{\mbox{\textsc{RCAN}}\xspace}
\newcommand{\ESRGAN}{\mbox{\textsc{ESRGAN}}\xspace}

\newcommand{\HVAE}{\mbox{\textsc{LVAE}}\xspace}

\newcommand{\UNet}{\mbox{\textsc{U-Net}}\xspace}

\newcommand{\figTeaserMain}{
\begin{figure}[t!]
  \centering
    \includegraphics[width=.99\linewidth]{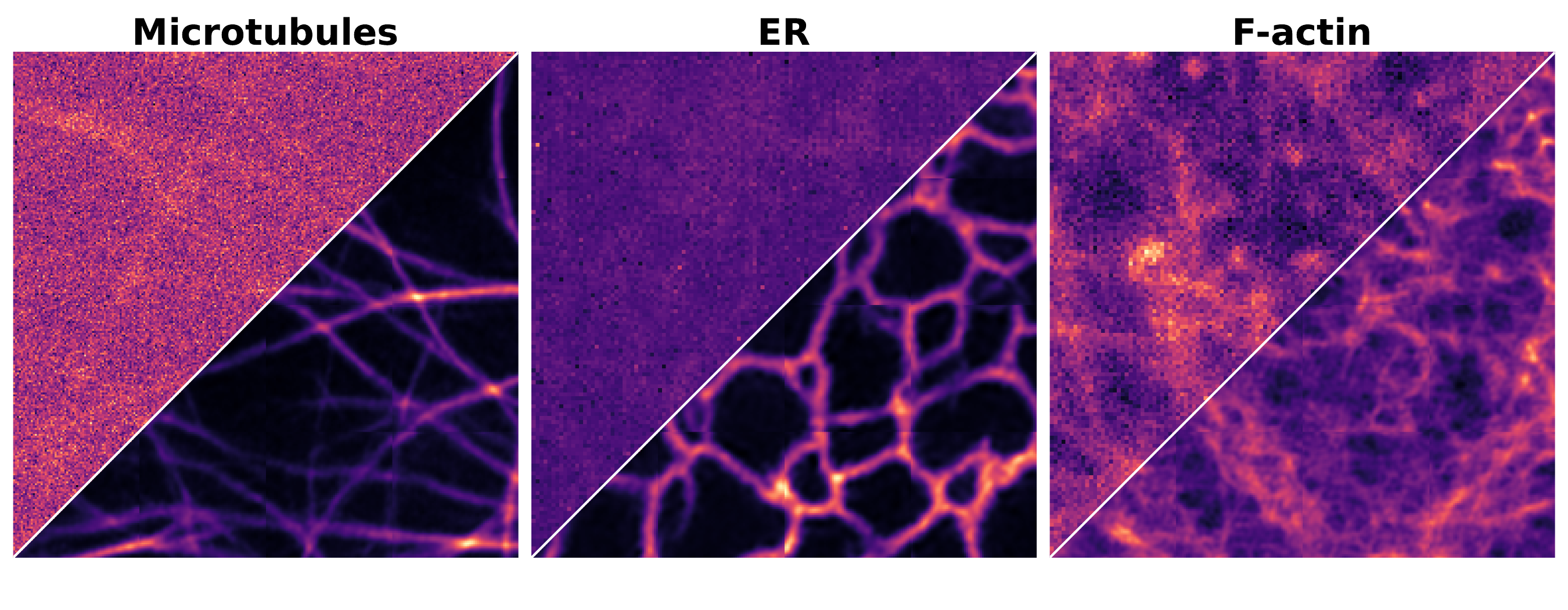}
    \vspace{-5mm}
    \caption{
    \ResM, using a conditional flow matching approach, leads to best-in-class computational super-resolution results even in severely noisy microscopy data.}
  \label{fig:teaser}
\end{figure}
}

\newcommand{\figPlotsMain}{
\begin{figure*}[ht!]
\centering
  \includegraphics[width=0.9\textwidth]{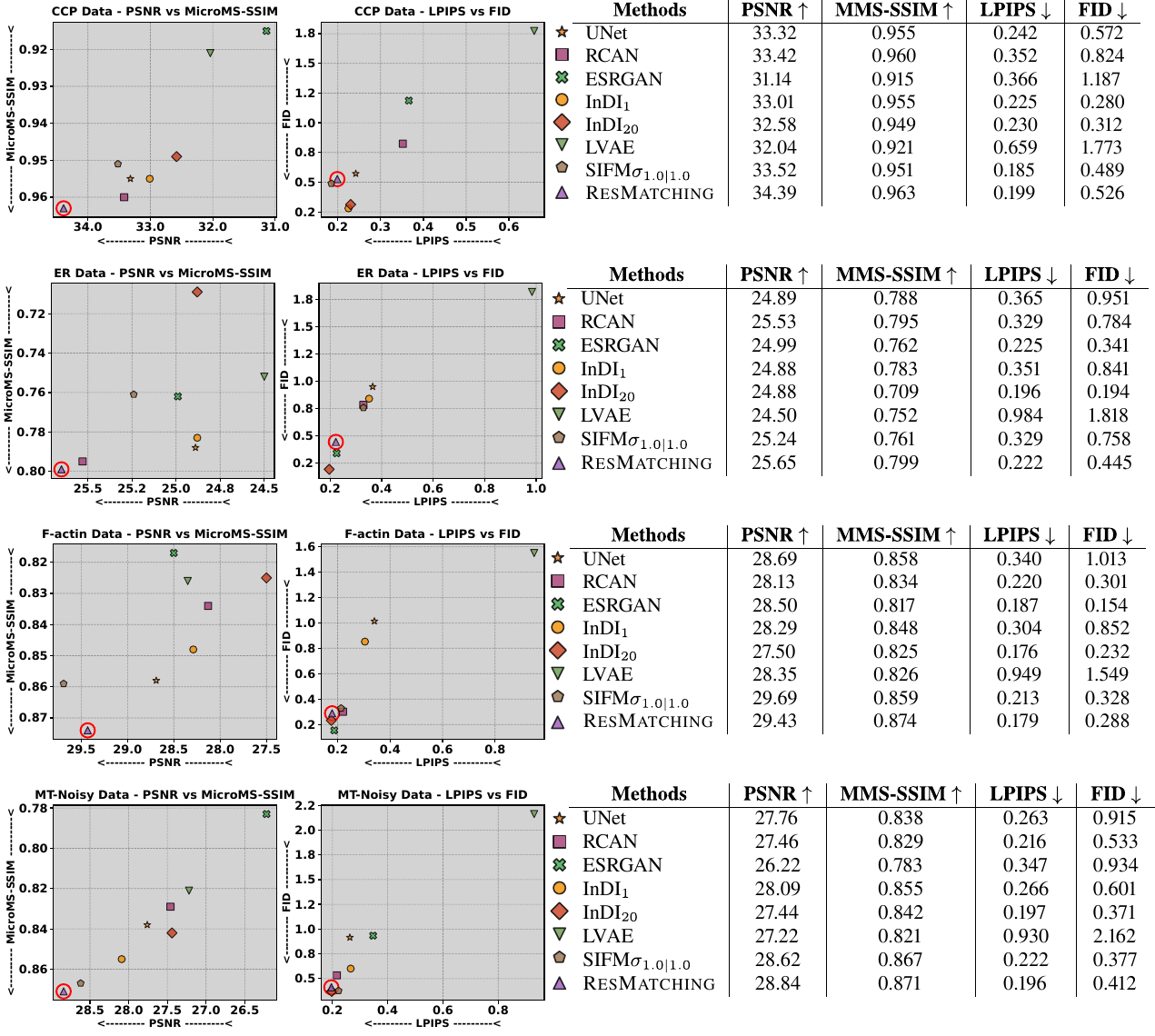}
\caption{
\textbf{Quantitative results -- data fidelity and realism.} 
Each row corresponds to results obtained on each data-subset we trained on (see Section~\ref{sec:experiments}).
We show PSNR vs.\ MicroMS-SSIM (left, note inverted axis for more consistent plots) and LPIPS vs.\ FID (center), capturing the trade-off between pixel-level fidelity and perceptual quality. 
\ResM is highlighted with an additional red circle. 
Results tables (right) show the same data. 
}
  \label{fig:plot_main}
\end{figure*}
}

\newcommand{\figQualitative}{
\begin{figure*}[ht!]
\centering
  \includegraphics[width=.99\linewidth]{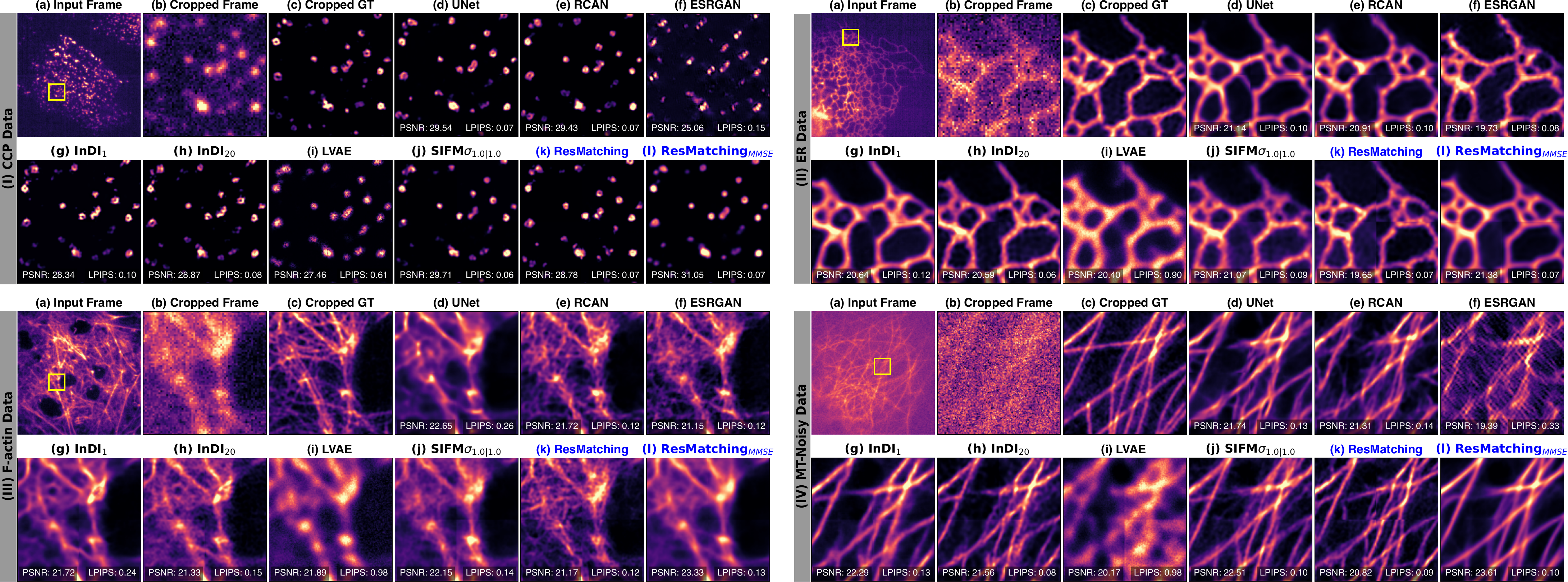} 
\caption{
\textbf{Qualitative results.} 
Representative predictions are shown for all data subsets (experiments). 
For each result \textbf{(I-IV)}, we show:
\textbf{(a)}~the full low-resolution noisy input with a highlighted crop region (yellow box), 
\textbf{(b)}~the cropped input region, 
\textbf{(c)}~the high-resolution ground truth, 
\textbf{(d–j)}~predictions of baseline methods (see Section~\ref{sec:experiments}), 
\textbf{(k)}~a single posterior sample generated by \ResM, and 
\textbf{(l)}~the MMSE estimate computed over 50 posterior samples. 
}
  \label{fig:qualitative}
\end{figure*}
}

\newcommand{\figCalibration}{
\begin{figure}[ht!]
\centering
  \includegraphics[width=0.77\linewidth]{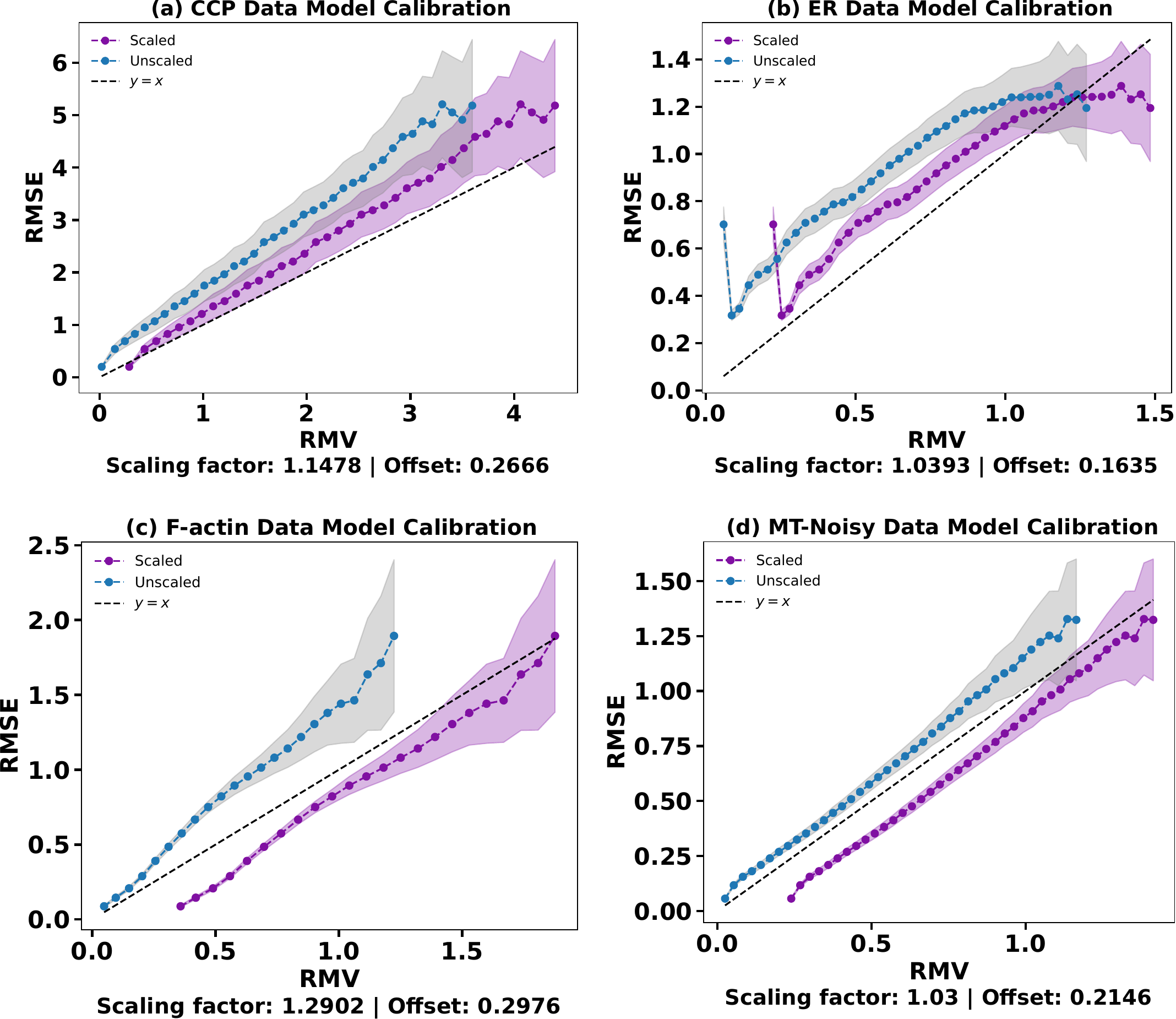} 
  \caption{\textbf{Model calibration.} 
  We show root mean variance~(RMV) versus root mean square error~(RMSE) as described in Section~\ref{sec:method}.
  Each plot ((a)--(d)) corresponds to one experiment we conducted.
  The dashed line indicates $y=x$.
}
  \label{fig:calib}
\end{figure}
}

\usepackage{fancyhdr}
\fancypagestyle{plain}{
  \fancyhf{}
  \fancyfoot[L]{arxiv.org}
  \fancyfoot[C]{\thepage}

}

\usepackage{orcidlink}


\title{ResMatching: Noise-Resilient Computational Super-Resolution\\ via Guided Conditional Flow Matching}

\name{
Anirban Ray$^{1,2}$,
Vera Galinova$^{1}$, and
Florian Jug$^{1}$
}
\address{
$^{1}$Human Technopole, Milan, Italy \quad
$^{2}$Technische Universität Dresden, Germany \\
{\tt\small \{anirban.ray, vera.galinova, florian.jug\}@fht.org}
}

\begin{document}
\ninept
\maketitle

\begin{abstract}
Computational Super-Resolution (CSR) in fluorescence microscopy has, despite being an ill-posed problem, a long history~\cite{Tian2025-uk}. 
At its very core, CSR is about finding a prior that can be used to extrapolate frequencies in a micrograph that have never been imaged by the image-generating microscope.
It stands to reason that, with the advent of better data-driven machine learning techniques, stronger prior can be learned and hence CSR can lead to better results.
Here, we present \textbf{\ResM}, a novel CSR method that uses guided conditional flow matching to learn such improved data-priors.
We evaluate \ResM on $4$ diverse biological structures from the \texttt{BioSR} dataset~\cite{biosr} and compare its results against $7$ baselines. 
\ResM consistently achieves competitive results, demonstrating in all cases the best trade-off between data fidelity and perceptual realism~\cite{PDT}.
We observe that CSR using \ResM is particularly effective in cases where a strong prior is hard to learn, \eg when the given low-resolution images contain a lot of noise.
Additionally, we show that \ResM can be used to sample from an implicitly learned posterior distribution and that this distribution is calibrated~\cite{calibration} for all tested use-cases, enabling our method to deliver a pixel-wise data-uncertainty term that can guide future users to reject uncertain predictions~\cite{hdn,usplit,denoisplit}. 
\end{abstract}

\begin{keywords}
computational super-resolution, conditional flow matching, uncertainty estimation.
\end{keywords}
\section{Introduction}
\label{sec:intro}

In the context of fluorescence microscopy, many approaches for computational super-resolution (CSR) have been proposed over the years~\cite{Tian2025-uk}. 
Classical CSR algorithms can be understood as attempts to invert the low-pass filtering imposed by the microscope's optical transfer function (OTF) and thereby to extrapolate high-frequency information lost during image formation. 
Because this inverse problem is ill-posed, priors are essential for constraining the space of feasible solutions~\cite{Tian2025-uk}. 
Early methods introduced explicitly designed priors that capture simple image characteristics, such as smoothness or sparsity. 
Prominent examples include Tikhonov~\cite{tikhonov} and total-variation regularization~\cite{tv}, which penalize high gradients to suppress noise while preserving edges, and iterative deconvolution schemes such as the Richardson–Lucy algorithm~\cite{RL}, which exploit a Poisson noise model and an implicit positivity constraint to recover plausible high-frequency detail.

\figTeaserMain

With the advent of deep learning, hand-crafted regularization terms have been replaced by learned, data-driven priors. 
Early convolutional architectures such as the \UNet~\cite{unet,CARE} and \RCAN~\cite{rcan} established strong baselines for microscopy restoration by learning hierarchical feature representations and attention-based refinement, respectively. 
However, their deterministic nature limits their ability to model the inherent uncertainty in fluorescence imaging, especially under high noise conditions.
These models, trained directly on suitable microscopy datasets, capture the statistics of biological structures directly from the data and enable more powerful solutions to inverse problems such as denoising, deblurring, or super-resolution~\cite{floriansReviewWithSulianaEtAl}. 
In particular, convolutional and adversarial networks~\cite{esrgan} have demonstrated that structural priors learned from fluorescence micrographs can substantially improve resolution restoration, even under severe photon limitations. 
From this perspective, modern CSR can be viewed as the process of learning a {\em structural prior} that constrains the predictor toward biologically meaningful high-frequency content. 

To overcome these limitations, probabilistic frameworks such as ladder variational autoencoders (\HVAE~\cite{lvae}) and implicit diffusion inference models~\cite{con_diff} have introduced stochastic latent variables to capture spatial uncertainty and diverse, plausible reconstructions.
Recent progress in generative modeling has further expanded this view. 
Models such as variational autoencoders~\cite{lvae}, diffusion processes (\InDI~\cite{indi}), and flow matching networks~\cite{sifm, hazematching} provide a formal framework to approximate the distribution of high-resolution images conditioned on low-resolution measurements. 
Such generative approaches also offer a flexible way to sample from an explicit or implicit posterior over possible reconstructions, which, in theory, enables uncertainty quantification.

In this work, we investigate whether conditional flow matching, a recent generative modeling paradigm, can yield improved priors for CSR in fluorescence microscopy. 
Rather than aiming for a general-purpose foundation model, we introduce \textbf{\ResM}, which 
$(i)$~learns a detailed, data-specific structural prior from fluorescence microscopy images, 
$(ii)$~achieves a good trade-off between data fidelity and perceptual realism~\cite{PDT}, 
$(iii)$~supports posterior sampling to visualize the diversity of plausible reconstructions, and 
$(iv)$~provides well-calibrated uncertainty estimates~\cite{calibration}. 
Together, these features make \ResM a conceptually novel and practically powerful method for uncertainty-aware computational super-resolution.

\section{Method}
\label{sec:method}

\subsection{Problem definition}
Computational super-resolution (CSR) in fluorescence microscopy seeks to recover high-resolution (HR) structures from low-resolution (LR) acquisitions corrupted by optical blur and imaging noise. 
We model the LR image as $\mathbf{x}_{M_0} = H_{M_0}(s) + \eta(s)$, where $H_{M_0}$ represents the unknown degradation operator of microscope $M_0$, $s$ is the underlying biological structure, and $\eta(s)$ is the signal-dependent noise. 
The corresponding HR image from a higher resolution microscope $M_1$ is given by $\mathbf{x}_{M_1} = H_{M_1}(s) + \eta(s)$. 
Given paired LR–HR observations $(\mathbf{x}_{M_0}, \mathbf{x}_{M_1})$ of the same sample $s_i \in \mathcal{S}$, our goal is to learn a function $\mathbf{\hat{x}}_{M_1} = S(\mathbf{x}_{M_0})$ that reconstructs HR-like images $\mathbf{\hat{x}}_{M_1}$ that are faithful to the unknown $\mathbf{x}_{M_1}$.

\subsection{Conditional flow matching (CFM)}
Flow matching unifies diffusion processes and normalizing flows through continuous-time dynamics~\cite{sifm,hazematching}. 
Instead of incrementally denoising a sample through discrete diffusion steps, flow-matching models learn a vector field that continuously transports samples from a simple base distribution to the target data distribution. 
In its conditional form, the model learns a velocity field $v_{\theta}(t, \mathbf{x}_t, \mathbf{x}_{M_0})$ such that a clean HR image $\mathbf{x}_{M_1}$ can be recovered from its degraded counterpart $\mathbf{x}_{M_0}$ by integrating the associated ODE, $\frac{d \mathbf{x}_t}{d t} = v_{\theta}(t,\mathbf{x}_t, \mathbf{x}_{M_0})$, with $\mathbf{x}_0 = \mathbf{x}_{t=0} \sim \mathcal{N}(0,I)$ and $\mathbf{\hat{x}}_{M_1} = \mathbf{x}_{t=1}$ corresponding to the reconstructed HR image.

\figPlotsMain

In fluorescence microscopy, the degradation operator $H_{M_0}$ is typically unknown and highly variable across microscopes, while the acquisition conditions and noise statistics $\eta(s)$ depend on the true signal intensity. 
Standard conditional flow-matching methods~\cite{sifm} assume that $H_{M_0}$ and the noise level are known, an unrealistic assumption in our setting. 
Recent work such as \DeMicFlow~\cite{hazematching} addressed this limitation by introducing a guided variant of CFM that implicitly learns to invert complex degradations without access to $H_{M_0}$ or explicit noise parameters.

{\scriptsize
\begin{algorithm}
\caption{\ResM Training Procedure}\label{alg:train}
\begin{algorithmic}[1]
\Require velocity model $v_\theta$; 
learning rate $\gamma$;
steps $T$
\Repeat
\State $s \sim \mathcal{S}$; 
$t \sim \mathcal{U}([i/T]_{i=0}^{T})$; 
$\mathbf{x}_0 \sim \mathcal{N}(0,I)$;
\State $\mathbf{x}_{M_0} \sim p_{M_0}(\mathbf{x}|s)$; $\mathbf{x}_{M_1} \sim p_{M_1}(\mathbf{x}|s)$
\State $\mathbf{x}_t=(1-t)\mathbf{x}_0+t\mathbf{x}_{M_1}$
\State $\mathcal{L}=||v_\theta(t,\mathbf{x}_t,\mathbf{x}_{M_0})-(\mathbf{x}_{M_1}-\mathbf{x}_0)||^2$
\State $\theta \leftarrow \theta-\gamma\nabla_\theta\mathcal{L}$
\Until{converged}
\State \Return $v_\theta$
\end{algorithmic}
\end{algorithm}
}


{\scriptsize
\begin{algorithm}
\caption{\ResM Inference Procedure}\label{alg:infer}
\begin{algorithmic}[1]
\Require velocity model $v_\theta$, steps $T$, input $\mathbf{x}_{M_0}$ to condition on
\State $\delta = 1/T$
\State $\mathbf{x}_0 \sim \mathcal{N}(0,I)$
\For{$t=1$ to $T$}
\State $\mathbf{x}_t = \mathbf{x}_{t-1} + \delta \, v_\theta(\delta\cdot(t-1),\mathbf{x}_t,\mathbf{x}_{M_0})$
\EndFor
\State \Return $\mathbf{x}_T$ 
\textcolor{gray}{\Comment{called $\hat{\mathbf{x}}_{M_1}$ in Section~\ref{sec:method}}}
\end{algorithmic}
\end{algorithm}
}

\subsection{Guided conditional flow matching for CSR}
To learn $S$, we extend the conditional flow matching framework proposed in \DeMicFlow~\cite{hazematching} to the CSR setting by guiding the conditional velocity field with the observed LR image $\mathbf{x}_{M_0}$. 
The model learns a continuous transport that maps samples drawn from a base Gaussian distribution to samples from the HR image distribution, conditioned on the LR input.
This yields a generative formulation in which HR reconstructions are obtained by integrating the learned conditional velocity field from $t=0$ to $t=1$.

\subsection{Marginal probability path}
We consider three relevant distributions: a base distribution $\mathbf{x}_0 \sim \mathcal{N}(0, I)$, the LR (source) distribution $\mathbf{x}_{M_0} \sim p_{M_0}(\mathbf{x}|s_i)$, and the HR (target) distribution $\mathbf{x}_{M_1} \sim p_{M_1}(\mathbf{x}|s_i)$. 
We define the interpolant between $\mathbf{x}_0$ and $\mathbf{x}_{M_1}$ as $\mathbf{x}_t = (1-t)\mathbf{x}_0 + t \mathbf{x}_{M_1}$ for $t \in [0,1]$, leading to the conditional path distribution 
\begin{equation}
p_t(\mathbf{x}|\mathbf{x}_{M_1}) = \mathcal{N}(t \mathbf{x}_{M_1}, (1-t)^2 I).
\end{equation}

\subsection{Marginal velocity field for CSR}
Our velocity field $v(t,\mathbf{x}_t, \mathbf{x}_{M_0}|\mathbf{x}_{M_1})$ depends on the interpolation time step $t\in[0,1]$, the interpolant $\mathbf{x}_t$, and the guiding LR image $\mathbf{x}_{M_0}|\mathbf{x}_{M_1}$ on which we condition. 
The marginal velocity field is then defined as
\begin{equation}
v(t,\mathbf{x}_t, \mathbf{x}_{M_0}) = \int v(t, \mathbf{x}_t|\mathbf{x}_{M_1}, \mathbf{x}_{M_0}) \, p_{M_1}(\mathbf{x}_{M_1}|\mathbf{x}_t, \mathbf{x}_{M_0}) \, d\mathbf{x}_{M_1}.
\end{equation}
We train a neural network $v_{\theta}$ to approximate this velocity field by minimizing
\begin{equation}
\min_{\theta}\mathbb{E}_{(\mathbf{x}_0, \mathbf{x}_{M_0}, \mathbf{x}_{M_1}), t}
\left\| v_{\theta}(t,\mathbf{x}_t, \mathbf{x}_{M_0}) - (\mathbf{x}_{M_1} - \mathbf{x}_0) \right\|^2,
\end{equation}
where $(\mathbf{x}_0, \mathbf{x}_{M_0}, \mathbf{x}_{M_1}) \sim p(\mathbf{x}_0, \mathbf{x}_{M_0}, \mathbf{x}_{M_1}|s_i)$ and $\mathbf{x}_t \sim p_t(\mathbf{x}|\mathbf{x}_{M_1})$. 
The network learns to predict the conditional velocity that transports a noisy sample $\mathbf{x}_0$ toward $\mathbf{x}_{M_1}$, guided by the LR input $\mathbf{x}_{M_0}$. 

\subsection{Inference and posterior sampling}
At inference time, the CSR operator $S(\mathbf{x}_{M_0})$ integrates the learned velocity field from $t=0$ to $1$ via an ODE solver,
\begin{equation}
\frac{d\mathbf{x}_t}{dt} = v_{\theta}(t,\mathbf{x}_t, \mathbf{x}_{M_0}), \qquad \mathbf{x}_0 \sim \mathcal{N}(0,I).
\end{equation}
The final state $\mathbf{x}_1$ corresponds to the CSR prediction $\mathbf{\hat{x}}_{M_1}$. 
Multiple predictions from an implicit posterior are possible by repeating the inference from different $\mathbf{x}_0^j \sim \mathcal{N}(0,I)$. 
Details of training and inference are similar to what was discussed in~\cite{hazematching} and are shown here in detail in Algorithms~\ref{alg:train} and~\ref{alg:infer}, respectively.

\subsection{Uncertainty calibration}
Being capable of sampling solutions from an implicit posterior, see previous paragraph, we can adopt the uncertainty calibration strategy from~\cite{denoisplit,Microsplit,hazematching}. 
Once posterior samples $\{\mathbf{\hat{x}}^{(k)}_{M_1}\}_{k=1}^K$ were generated for all $K$ LR input images, we computed the pixel-wise standard deviation over those samples per image and combined all these values to estimate the model uncertainty via the root mean variance (RMV). 
We additionally fit a linear calibration model between the RMV and the true prediction error, quantified via the root mean squared error (RMSE) \wrt a required set of ground truth HR images, by minimizing
$\min_{\alpha, \beta} \| \text{RMSE} - (\alpha \, \text{RMV} + \beta) \|^2$,
and report both calibrated and uncalibrated reliability plots to evaluate prediction trustworthiness (see Figure~\ref{fig:calib}).

\figQualitative
\figCalibration

\section{Experiments}
\label{sec:experiments}

\textbf{Datasets:} We evaluate \ResM on four representative biological structures from the \texttt{BioSR} dataset: Clathrin-Coated Pits (CCP), Endoplasmic Reticulum (ER), F-actin, and Microtubule-Noisy (MT-Noisy), where we add additional noise to \texttt{BioSR's} MT data.
The training sets contain 39, 53, 35, and 40 raw images of size $1004 \times 1004$, from which we crop 3120, 4240, 2800, and 3200 $128{\times}128$ patches, respectively. 
Each training data subset and hence experiment further includes 5 validation images and 10 test images.

\textbf{Training and Evaluation:} We train our networks according to~\cite{hazematching}, taking into account the adaptations for the CSR task we described in Section~\ref{sec:method} and Algorithm~\ref{alg:train} using a step size according to $T=20$. 
Note that we feed the interpolation time step $t$ to the velocity field $v_\theta$, after positionally encoding it, to all residual blocks by adding it to their respective output tensor~\cite{diffusinOpenAI}.
After training, to evaluate the quality of the trained models, each test image is processed according to Algorithm~\ref{alg:infer}, using inner tiling with 50\% overlap~\cite{usplit}, \ie from each $128{\times}128$ tile only the central $64{\times}64$ region is used to stitch the final prediction.
Distortion metrics (PSNR, MicroMS-SSIM~\cite{microssim}) are then computed over the entire stitched image.
Because perceptual metrics (LPIPS, FID) are sensitive to tiling boundaries, we compute them on individual $64{\times}64$ inner tiling crops.
Inference for a full test image requires $3.976\pm 0.003$\,sec.\ on an NVIDIA V100 GPU.

\textbf{MMSE estimate:} Since \ResM can sample from an implicit posterior, we also compute an approximate MMSE estimate by pixel-wise averaging $50$ posterior samples~\cite{hdn}.

\textbf{Baselines:} We compare \ResM against point-predicting and variational/generative baselines.
As point predictors, we chose a vanilla \UNet~\cite{unet} and an \RCAN~\cite{rcan}. 
Among generative approaches, we evaluate against the adversarially trained \ESRGAN~\cite{esrgan}, \InDI~\cite{indi} (both with 1-step and 20-step inference), and a ladder variational autoencoder (\HVAE)~\cite{lvae}. 
We also compare against SIFM~\cite{sifm}, setting $\sigma{=}1.0$ following the optimal configuration reported in~\cite{hazematching}.
\section{Results}
\label{sec:results}

Figure~\ref{fig:plot_main} shows the PSNR--MicroMS-SSIM and LPIPS--FID trade-offs for each training set described above. 
Across all experiments, \ResM achieves competitive fidelity scores (PSNR, MicroMS-SSIM) while simultaneously maintaining very good perceptual quality (LPIPS, FID). 
These results demonstrate that our conditional flow-matching approach learns a structural prior strong enough to outperform all baselines.
As indicated in~\cite{hazematching,indi}, the perception–distortion trade-off can additionally be modulated by varying the $T$ during inference.

In Figure~\ref{fig:qualitative}, we show representative crops for each dataset, comparing all methods with one another.
Note that for \ResM we show a single posterior sample (prediction) as well as the approximate MMSE.
While baselines such as \RCAN and \ESRGAN tend to over-smooth or hallucinate high-frequency details, \ResM preserves filament continuity and realistic texture. 

\textbf{Model Calibration:} We plot the root-mean-variance (RMV) against the root-mean-squared-error (RMSE) for all datasets (Fig.~\ref{fig:calib}). 
After linear rescaling, the calibration curves remain close to the ideal identity line, indicating that all trained networks are well calibrated~\cite{calibration}.

\section{Discussion and Conclusion}
\label{sec:disc_conc}

Computational super-resolution (CSR) ultimately depends on the strength and validity of the prior used to infer frequencies that were never measured by the microscope. 
\ResM contributes to this ongoing pursuit by demonstrating that guided conditional flow matching can learn expressive, biologically grounded priors that unify denoising and resolution enhancement in a single generative process. 
At the same time, our findings highlight the intrinsic difficulty of CSR: predicting unseen high-frequency structures is fundamentally uncertain, and visually plausible results do not necessarily imply physical correctness~\cite{floriansReviewWithSulianaEtAl,Tian2025-uk}. 
By explicitly modeling uncertainty and calibrating the learned posterior, \ResM offers a principled way to quantify and communicate this ambiguity. 

We hope that this work not only advances the technical state of CSR but also encourages a more careful interpretation of its outputs—treating super-resolved reconstructions as informed hypotheses rather than as ground truth. 
\section{Compliance with ethical standards}
\label{sec:ethics}
This research was conducted using publicly available microscopy datasets that do not contain any human or animal subjects. No ethical approval was required.

\section{Acknowledgments}
\label{sec:acknowledgments}

This work was supported by the European Union through the Horizon Europe program (IMAGINE project, grant agreement 101094250-IMAGINE and AI4Life project, grant agreement 101057970-AI4LIFE) and the generous core funding of Human Technopole. We also thank Ashesh for the fruitful discussions.

\bibliographystyle{IEEEbib}
\bibliography{strings,refs}

\end{document}